# Terrain Inclination Aided Three-Dimensional Localization and Mapping for an Outdoor Mobile Robot

Regular Paper

Xiaorui Zhu[1,*], Chunxin Qiu[1] and Mark A. Minor[2]

1 State Key Laboratory of Robotics and System (HIT), Harbin Institute of Technology Shenzhen Graduate School, Shenzhen, China
2 Department of Mechanical Engineering, University of Utah, Salt Lake City, USA
* Corresponding author E-mail: xiaoruizhu@hitsz.edu.cn

Abstract A new 3D localization and mapping technique with terrain inclination assistance is proposed in this paper to allow a robot to identify its location and build a global map in an outdoor environment. The Iterative Closest Points (ICP) algorithm and terrain inclination-based localization are combined together to achieve accurate and fast localization and mapping. Inclinations of the terrains the robot navigates are used to achieve local localization during the interval between two laser scans. Using the results of the above localization as the initial condition, the ICP algorithm is then applied to align the overlapped laser scan maps to update the overhanging obstacles for building a global map of the surrounding area. Comprehensive experiments were carried out for the validation of the proposed 3D localization and mapping technique. The experimental results show that the proposed technique could reduce time consumption and improve the accuracy of the performance.

Keywords Mobile Robot, Localization, Mapping, Terrain Inclinations

## 1. Introduction

Laser range sensors have been widely used in outdoor environments to achieve localization and mapping because they can provide accurate and ample point clouds of the surrounding environment. However, the computational cost of dealing with point clouds is very expensive, especially for poorly structured outdoor scenarios, which results in large time-consumption in relation to task execution. Therefore, the goal of this research is to provide a less computationally-demanding and more accurate 3D localization and mapping approach with terrain inclination assistance for the mobile robot navigating on inclined terrains.

Mobile robots have been increasingly widely applied in many different scenarios, such as space exploration [1] and search and rescue [2], where the robots are required to travel over uneven terrain while outdoors. Recently, many aspects of research on mobile robots operating in uneven terrains have been studied, such as rough terrain modelling [3, 4], terrain characterization [5, 6] and motion planning and control [4]. Autonomous localization and mapping is one of the vital and challenging problems for outdoor mobile robots.

In the last decade, significant investigation of mobile robot localization and mapping has been performed [7]. The simultaneous localization and mapping (SLAM) was proven to be effective in generating large, consistent maps, and achieving localization, especially for indoor applications [8, 9]. Different sensors have been used to obtain the surrounding information, such as laser range finders [8, 10, 11], vision systems [9, 12, 13] and sonar [14, 15]. These sensors are used to generate a serial of local maps as the robot is moving. Such overlapped local maps need to be matched so as to obtain the relative rotation and translation for localization and in order to build a global map of the whole navigation area [16]. Among these sensors, laser range sensors are more popular because they can provide more reliable perception and a wider view of the surrounding environment in facilitating localization and mapping. Some researchers have proposed the combination of a 3D laser range scanner with a 2D simultaneous localization and mapping algorithm for autonomous navigation in uneven environments [10, 11]. Hence, this paper also takes advantage of a 3D laser range scan system.

Traditionally, the Iterative Closest Points (ICP) algorithm [17] is used to align the overlapped local maps, whereby the overhanging obstacles are generally treated as landmarks [10]. ICP has been proven to be very effective in building the global map of the surrounding area by dealing with serial overlapped local maps. The inherited disadvantages of ICP, however, are slow convergence and locally optimal solutions. In order to overcome these shortcomings, some researchers have developed different variations of ICP algorithms [10, 18]. Notice that it is much more challenging to achieve efficient localization and mapping in most outdoor environments. Sparse or unconstructed overhanging obstacles can greatly harm precision and complex point clouds data can increase the time-consumption of ICP algorithms. Hence, Kummerle et al. proposed the outdoor localization and mapping assisted by terrain with a multilevel surface [19]. Lee et al. have made an effort to improve the performance of ICP-based localization and mapping using terrain classification [16]. In unstructured outdoor environments, even one ICP matching was very time-consuming (more than 30s) [20]. Later, Nüchter et al. proposed an approximate kd-tree which reduced the time consumption of each ICP matching to roughly 75% [21]. Furthermore, Nüchter et al. proposed a new approach called the 'cached kd-tree' for large environments and compared the proposed cached kd-tree with several different methods in terms of their computational costs, such as point reduction and kd-tree, while the average distance between two consecutive 3D scans was 2.5 m [22]. They found out in [22] that each ICP matching took one second with cached kd-tree after point reduction. As is currently known, this is the fastest ICP matching for unstructured outdoor applications, although the time consumption of each ICP matching could be milliseconds for many indoor scenarios. However, if more accurate localization and mapping results are needed, ICP matching has to be applied for more instances, such that the computation cost becomes much more expensive [23]. Therefore, frequent ICP matching still results in expensive computational costs, although the matching time for each alignment can be reduced to a reasonable range. Moreover, the consumption of each scan is not negligible (usually more than 3 seconds) [21, 22]. Hence, the substantial time consumption of scans can also increase the total task execution period, especially while exploring large scale environments.

In order to balance the accuracy and computational load as well as to take advantage of ICP matching, a new method is proposed in this paper to combine the ICP algorithm and the inclination of the terrains the robot is navigating so as to achieve autonomous 3D localization and mapping. Using this method, fewer local maps are required to be generated by the 3D laser scan system. Hence, the time-consuming ICP matching is ultimately applied far fewer times in building a joint global map. Terrain inclinations are used to achieve local localization based on the ground points extracted from the previous laser scan data during the interval between two laser scans. The idea of this terrain inclination-based localization has been presented by our research group and has proven to be accurate and fast [24]. The time consumption is therefore reduced by preventing frequent scan-and-alignments of large point clouds.

The main contribution of this research involves a terrain inclination aided 3D localization and mapping technique (TILAM) to allow the robot to identify its position and build the global map of the navigation area. Experiments then verify the proposed approach and demonstrate accurate and fast localization and mapping on outdoor inclined terrains.

The structure of this paper is as follows. The laser scan data process is presented in Section 2. The terrain inclination aided localization and mapping technique is proposed in Section 3. Experimental results and a discussion are presented in Section 4. Conclusions are described in Section 5.

## 2. Laser Scan Data

*2.1 Data acquisition*

In this paper, a 3D laser scan system was built on the top of the mobile robot to obtain the point clouds of the environment during the robot navigation, Figure 1. The 3D laser scan system consists of a 2D laser scanner (SICK LMS200) and a turntable driven by a servo motor.

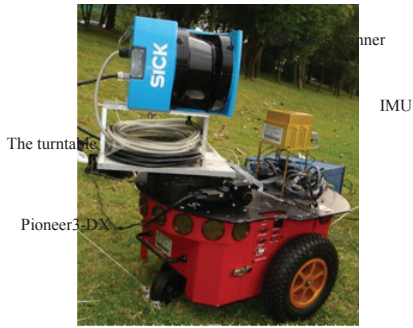

**Figure 1.** Experiment platform.

Three coordinate systems are defined in this paper. They are the world coordinate system $(x,y,z)$, the robot coordinate system $(x_b, y_b, z_b)$ and the laser coordinate system $(x_l, y_l, z_l)$. The world coordinate system is fixed to the tangential plane of the earth-surface-ellipsoid at the estimated initial position of the robot with three orthogonal axes $(x, y, z)$. The axis $y$ always points to the north-pole of the earth and the axis $z$ points upwards, away from the earth's centre. The robot coordinate system is fixed to the midpoint of the rear axis of the robot with three orthogonal axes $(x_b, y_b, z_b)$. The axis $x_b$ is always consistent with the direction of the robot velocity while the axis $z_b$ is vertical to the robot plane. The laser coordinate system is fixed to the mirror wheel centre of the laser scanner. The direction of the axis $x_l$ is always the same as that of the axis $x_b$ while the axis $z_l$ is coincident with the axis $z_b$.

In this paper, scanning the environment with a mobile robot is done in a stop-scan-go fashion. The laser scan is applied every a few meters. After the robot stops, the laser scanner is rotated by the turntable from 0º (positive direction of the $x_l$ axis) to 180º (negative direction of the $x_l$ axis). Hence, we can get the 3D coordinate values of each point clouds in the laser coordinate system as:

$$z_l = R \times \sin\gamma \quad (1)$$

$$y_l = R \times \cos\gamma \times \cos\varphi \quad (2)$$

$$x_l = R \times \cos\gamma \times \sin\varphi \quad (3)$$

where $(x_l, y_l, z_l)$ are the coordinate values, R is the distance from the laser to the object, $\gamma$ is the angle between the scan line and the $x_l o_l y_l$ plane and $\varphi$ is the rotation angle of the turntable.

*2.2 Point Clouds Separation*

The point clouds representing the environmental information contain both the terrain points and non-terrain points (overhanging points). Rather than the traditional approaches, the terrain points and non-terrain points both contribute to the localization and mapping in this research. Hence, those points need to be extracted separately from the point clouds map.

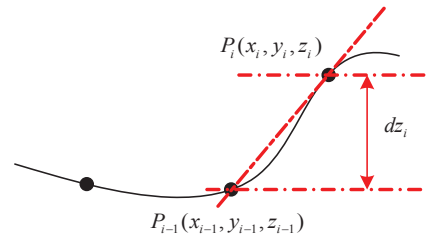

**Figure 2.** Relationship between two points.

In order to separate this point data, a modified filter is applied on each scan line based on so-called "pseudo scanning lines" by slopes and elevations [25]. Since a scan line is composed of many points ordered by the scanning sequence, we can arbitrarily choose two adjacent points $P_{i-1}$ and $P_i$. Then, $dz_i$ and $S_i$ represent the elevation difference and the slope between the points $P_i$ and $P_{i-1}$. If $dz_i$ and $S_i$ are both less than the threshold $Limit\_dz$ and $Limit\_S$, then the point $P_i$ is treated as the terrain point. Otherwise, it is a non-terrain point. According to Figure 2, the calculation formula of the elevation difference $dz_i$ and the slope $S_i$ are shown as:

$$dz_i = |z_i - z_{i-1}| \quad (4)$$

$$S_i = \frac{z_i - z_{i-1}}{\sqrt{(x_i - x_{i-1})^2 + (y_i - y_{i-1})^2}} \quad (5)$$

where $(x_i, y_i, z_i)$ is the world coordinate of the point $P_i$ and $(x_{i-1}, y_{i-1}, z_{i-1})$ is the world coordinate of the point $P_{i-1}$. Figure 3 shows the results of the data separation on a local point clouds scan where the green and red colours represent the terrain points and the non-terrain points respectively.

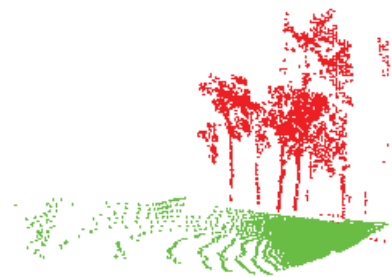

**Figure 3.** Data separation result.

### 3. Terrain Inclination Aided Localization and Mapping

*3.1 Terrain Inclination Aided Localization*

Assume a path is pre-planned such that the mobile robot can follow this path while achieving localization and mapping. The robot path is represented by the trajectory of the origin of the robot coordinate, $o_b$. The second assumption is that the inclined terrains where the robot might traverse have small elevation differences in their own neighbourhood patch. As we mentioned in the last section, the terrain points are used to help localization during the interval between two laser scans

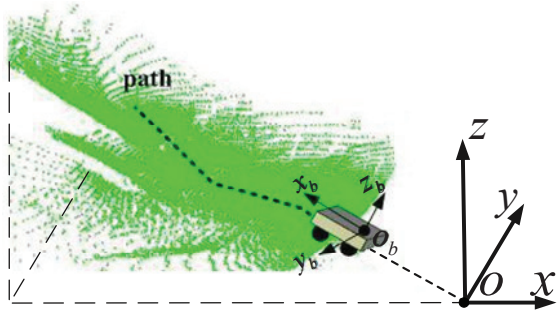

**Figure 4.** Point clouds map.

Figure 4 shows a 3D point clouds map from the laser scanner system captured at the $n^{th}$ scan. During the interval between the $n^{th}$ scan and the $(n+1)^{th}$ scan, the terrain inclinations will be correlated with the pitch and roll angles of the robot to localize the robot as it navigates on the inclined terrains. In order to find the relationship between the terrain inclinations and the robot attitude, the robot path is divided into many parts with a fixed length ($l$=15cm) that is equal to the length of the robot, Figure 5.

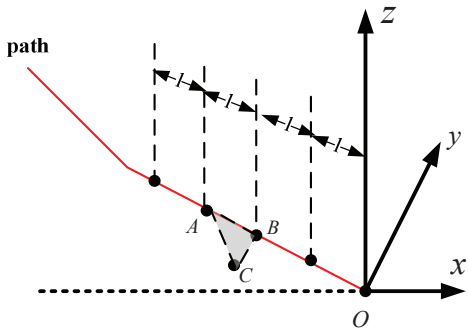

**Figure 5.** Extract the points from the cloud map.

In Figure 5, $A$ and $B$ are two points on the path whose distance is $l$ while the distance between the points $B$ and $C$ is half of the robot width. The line $BC$ is perpendicular to the line $AB$. Therefore, the triangle $ABC$ can be correlated with the robot plane at a specific time interval. Given a point clouds map generated by the laser scan system, the coordinate information of a series of triangular patches can be extracted across the path to represent the projection of the robot motion. Notice that all of this coordinate information is with respect to the world coordinate. Next, a robot-terrain inclination model $\{\gamma_M | \gamma_M = h_M(x,y,z)\}$ can be derived as a correlation function between the robot attitude and the robot position where $\gamma_M = \begin{bmatrix} \theta & \alpha & \phi \end{bmatrix}^T$ represents the orientation angles (yaw/pitch/roll) of the robot plane according to the terrain inclinations from the point clouds map. The heading angle, $\theta$, is exclusively determined by the path.

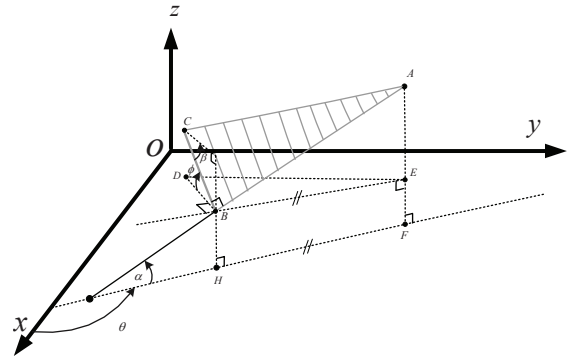

**Figure 6.** Geometric extraction of robot-terrain inclination model.

According to Figure 6, the angles, $\alpha$ and $\phi$ for one triangle are given by the following equations:

$$\alpha = \sin^{-1}(\frac{(z_3 - z_1)}{\overline{BA}}) \qquad (6)$$

$$\phi = \sin^{-1}(\frac{\sin\beta}{\cos\alpha}) \qquad (7)$$

$$\overline{BA} = \sqrt{(x_1 - x_3)^2 + (y_1 - y_3)^2 + (z_1 - z_3)^2} \qquad (8)$$

$$\overline{BC} = \sqrt{(x_1 - x_2)^2 + (y_1 - y_2)^2 + (z_1 - z_2)^2} \qquad (9)$$

where $B = (x_1, y_1, z_1), C = (x_2, y_2, z_2), A = (x_3, y_3, z_3)$ and $\beta = \sin^{-1}(\frac{z_2 - z_1}{\overline{BC}})$.

By extending the above method to the entire local point clouds map, a number of angles ($\theta_j, \alpha_j, \phi_j$) can be extracted from the serial triangular patches. Hence, we can get a series of discrete relationships between the robot attitude and the robot position. By fitting the above discrete relationship using linear regression, $\theta = h_1(x)$, $\alpha = h_2(y), \phi = h_3(z)$ can be obtained to constitute $h_M = \begin{bmatrix} h_1(x) & h_2(y) & h_3(z) \end{bmatrix}^T$.

Once we obtain this robot-terrain inclination model, an Extended Kalman Filter (EKF) is then applied to achieve localization. The state transition probability in one time interval is governed by:

$$\begin{bmatrix} x_t \\ y_t \\ z_t \\ \theta_t \\ \alpha_t \\ \phi_t \end{bmatrix} = \begin{bmatrix} x_{t-1} \\ y_{t-1} \\ z_{t-1} \\ \theta_{t-1} \\ \alpha_{t-1} \\ \phi_{t-1} \end{bmatrix} + \Delta t \begin{bmatrix} A_{3\times 3} & 0_{3\times 3} \\ 0_{3\times 3} & B_{3\times 3} \end{bmatrix} \begin{bmatrix} v_{x_b} \\ v_{y_b} \\ v_{z_b} \\ \omega_{x_b} \\ \omega_{y_b} \\ \omega_{z_b} \end{bmatrix} + \varepsilon_t \quad (10)$$

where:

$$A_{3\times 3} = \begin{bmatrix} c\theta c\alpha & -c\theta s\alpha s\phi - s\theta c\phi & -c\theta s\alpha c\phi + s\theta c\phi \\ s\theta c\alpha & -s\theta c\alpha s\phi + c\theta c\phi & -s\theta s\alpha c\phi - c\theta s\phi \\ s\alpha & c\alpha s\phi & c\alpha c\phi \end{bmatrix}_{t-1}$$

$$B_{3\times 3} = \begin{bmatrix} 1 & s\phi t\alpha & c\phi t\alpha \\ 0 & c\phi & -s\phi \\ 0 & s\phi/c\alpha & c\phi/c\alpha \end{bmatrix}_{t-1}.$$

The c, s and t in the above equations are the abbreviations for *cos*, *sin* and *tan* respectively. Δt is the sampling period. The linear velocity vector $\begin{bmatrix} v_{x_b} & v_{y_b} & v_{z_b} \end{bmatrix}$ and the angular velocity vector $\begin{bmatrix} \omega_{x_b} & \omega_{y_b} & \omega_{z_b} \end{bmatrix}$ are with respect to the body frame of the robot ($x_b$, $y_b$, $z_b$). $\varepsilon_t$ is the Gaussian random vector.

The robot-terrain inclination model extracted from the point clouds map is treated as the measurement probability function, Eq. (11),

$$\begin{bmatrix} \theta_t \\ \alpha_t \\ \phi_t \end{bmatrix} = \begin{bmatrix} h_1(x_t) \\ h_2(y_t) \\ h_3(z_t) \end{bmatrix} + \delta_t \quad (11)$$

where $\theta_t$, $\alpha_t$ and $\phi_t$ are the measurements of the yaw, pitch and roll angle of the robot from the on-board inertial sensors at the time interval *t* and $\delta_t$ is the Gaussian random vector.

*3.2 ICP-based Mapping*

In this paper, the ICP algorithm is used to align each pair of the local point clouds maps to finally build the global map of the navigation area. The ICP is based on a point-to-point correspondence between two partially overlapped scans. It starts with two meshes and an initial guess in order to get the relative transform between them. This is different from the traditional approach where the initial guess is given by odometry; the initial guess, in this paper, comes from the results of the terrain inclination aided localization applied in the previous time interval. Next, the initial transform is refined iteratively by repeatedly generating pairs of the corresponding points on the meshes and minimizing an error metric [26]. According to the literature [10], the ground information is not effective in achieving good matching in an outdoor environment, while the non-ground features like trees or buildings can get good and fast results. Point reduction and cached kd-tree are both applied here to reduce the time consumption of the ICP matching to around 1 second [22].

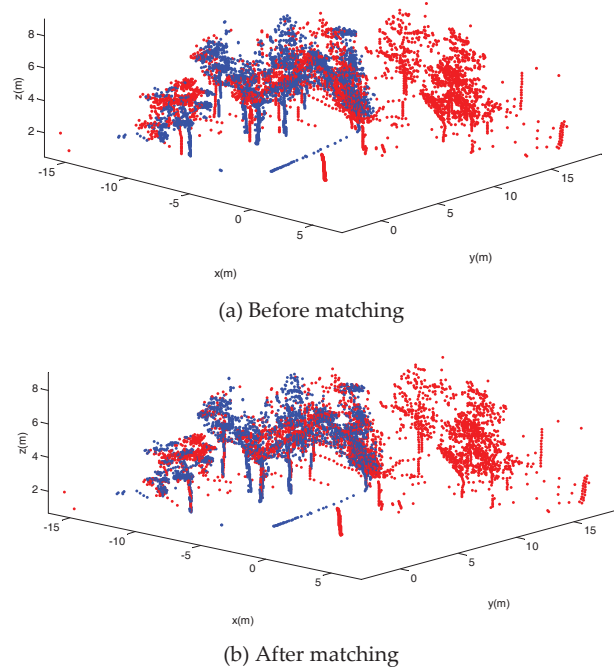

(a) Before matching

(b) After matching

**Figure 7.** Illustration of the alignment of two local point clouds maps

Figure 7 illustrates the matching of two local point clouds maps using non-terrain points and the initial guess from the terrain inclination aided localization. Figure 7 (a) shows two local point clouds maps before matching where the blue points represent the cloud map from the $n^{th}$ scan and the red points represent that from the $(n+1)^{th}$ scan. The matching results are shown in Figure 7 (b) after the alignment of two maps.

## 4. Experiments and Discussion

*4.1 Methods and Procedures*

The experiments were conducted on the platform, a Pioneer 3DX robot equipped with a 3D laser scan system, Figure 1. The NAV440 from Crossbow Technology® was used as the inertial measurement unit (IMU) that mounted on the robot in order to measure the roll, pitch, yaw angles and the angular velocities $\omega_{x_b}$, $\omega_{y_b}$, $\omega_{z_b}$. The forward speed $v_{x_b}$ with respect to the robot coordinate system was provided by the encoders of the robot. $v_{y_b}$, $v_{z_b}$ are both zero with respect to the robot coordinate system with no-slip assumption.

An outdoor environment is selected around the library of the Shenzhen University Town with the 3D terrains covered with the grass and surrounded with the trees. This environment is one kind of typical inclined terrains with natural landmarks (trees) where the proposed TILAM could be validated. This selected region is approximately 15m x 10m. The horizontal opening angle of the 3D laser scan system is 180 degrees while the vertical opening angle is 80 degrees. One scan map is obtained roughly every six meters for the proposed TILAM. In order to do a comparison, a fast ICP-based SLAM with point reduction and cached kd-tree [22] is applied at the same region where the robot stops and scans every three meters to update the estimation of the robot position. Note that this method is known as the fastest ICP variation in the literature so far. The robot moved on this terrain at a speed of 0.1m/s. The sampling period $\Delta t = 0.1s$ is chosen for all experiments so as to be consistent with the sampling period of IMU.

A PhaseSpace® motion capture system was applied here to obtain the reference positions of the robot during the navigation [27]. PhaseSpace® is an optical motion capture system to estimate the position, velocity and acceleration of an object to which LED markers are attached. Its measurement accuracy is 1 mm. Notice that the continuous ground truth value is difficult to obtain online since the motion capture system cannot cover the whole navigation area at one time. Hence, a few discrete artificial markers were labelled along the real robot path during each experiment. The corresponding time interval was also recorded for each marker. Next, the positions of those markers were measured individually by the motion capture system as the reference values after each experiment were done.

*4.2 Results and Discussion*

Figure 8 shows the robot-terrain inclination model extracted from the first scan map using the method in Section 3. Based on this robot-terrain model, the terrain inclination aided localization was achieved online during the interval between the first scan (*T*=0) and the second scan (*T*=1min), where *T* is the clock-time ticking from the beginning of the movement. At *T*=1min, the robot stopped and did the second scan. Afterwards, the ICP-based alignment between the first scan and the second scan was finished to generate a joined local map using the localization results at *T*=1min as the initial condition of the ICP, Figure 7. Likewise, the terrain inclination aided localization was applied again during the next interval between the second scan (*T*=1min) and the third scan (*T*=2min). A global map of this region was finally built after three scans while the online localization was also achieved during the whole navigation. The online localization results are shown in Figure 9 where the solid line represents the proposed TILAM, the dashed line represents the ICP-based SLAM, and the plus signs represent the reference positions. Figure 10 (b) and (c) show the mapping results of the region in Figure 10 (a) using the proposed TILAM and the ICP-based SLAM, respectively. Twelve discrete ground truth values were measured in total during each experimental trial as the references (the red flags in Figure 10 (a)).

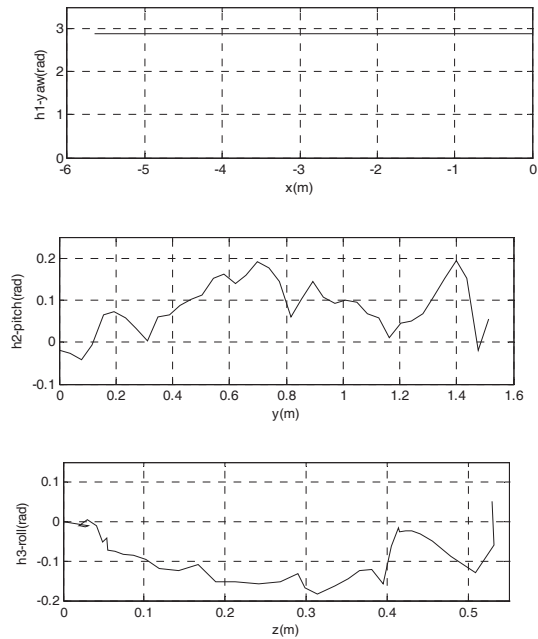

**Figure 8.** Robot-terrain inclination model based on the first scan map.

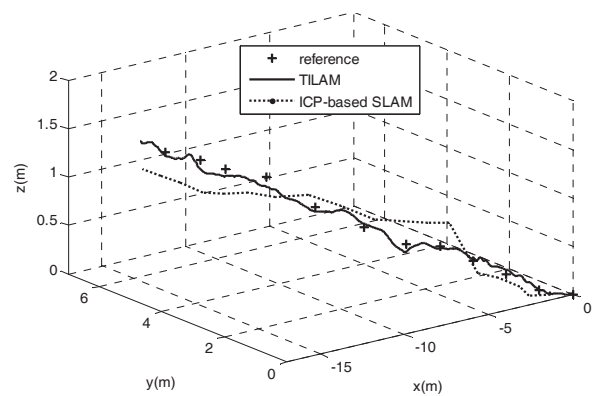

**Figure 9.** The localization result comparison.

According to Figure 10, there is no big difference between the proposed TILAM and the ICP-based SLAM in terms of the global map accuracy. Both of them can successfully build the global map. However, the position estimation by the proposed technique is much closer to the ground truth values than the traditional ICP-based SLAM in such an outdoor scenario as Figure 11. In order to further examine how different the localization performance between the proposed technique and the traditional ICP-based SLAM, five repeated experimental trials were carried out. The average mean and the variance of the position estimation errors in three individual axes and in the Euclidean distance, *d*, are shown, respectively, in Table 1. According to Table 1, the position estimation error in the Euclidean distance using the proposed TILAM is 82% smaller than that using the ICP-based SLAM. The distribution of the position errors from the TILAM is also 43% more narrow than the ICP-based SLAM.

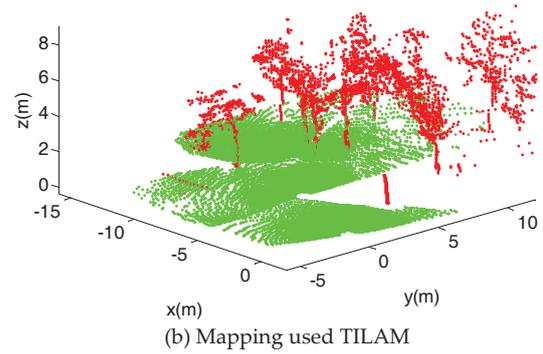

(b) Mapping used TILAM

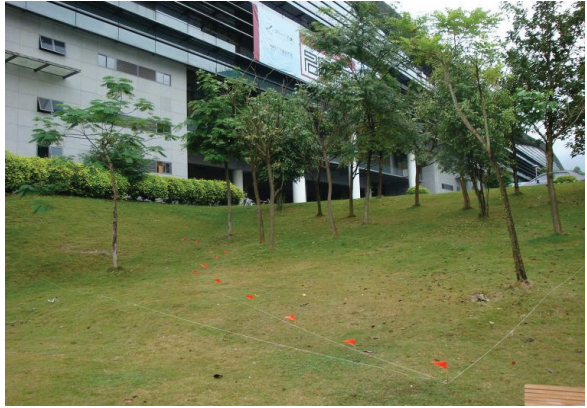

(a) Experimental environment

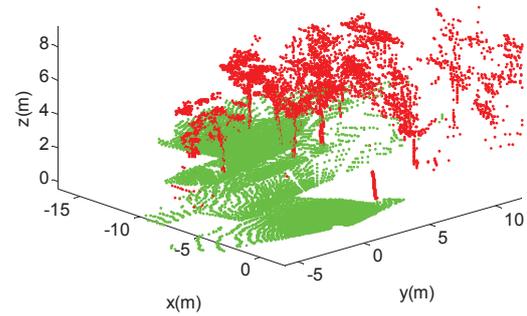

(c) Mapping used the ICP-based SLAM

**Figure 10.** The mapping result comparison.

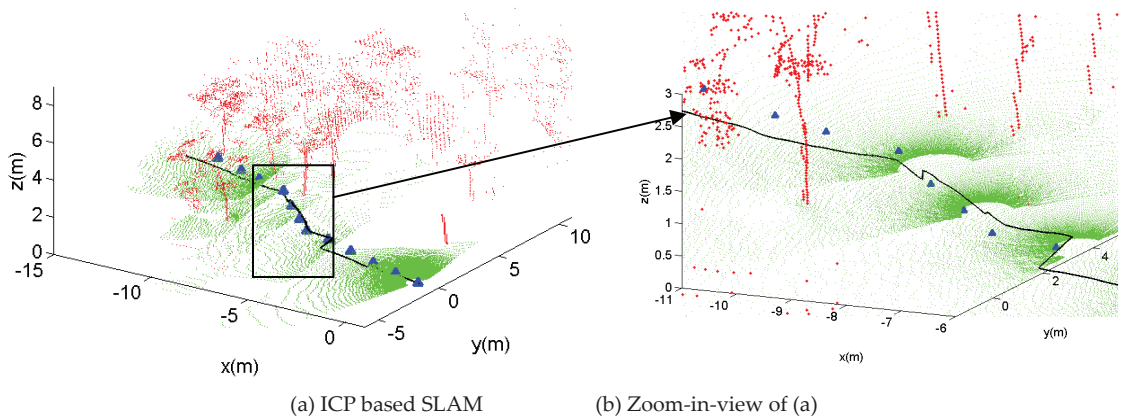

(a) ICP based SLAM      (b) Zoom-in-view of (a)

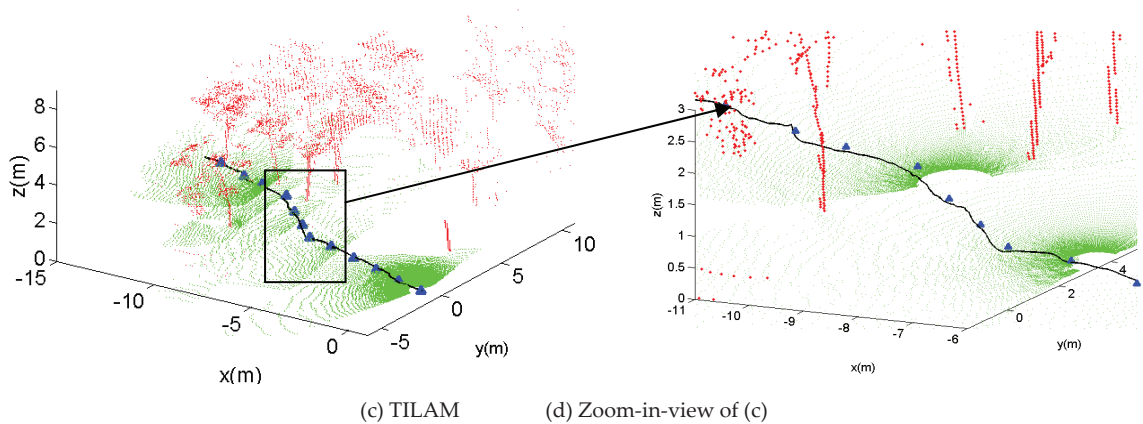

(c) TILAM      (d) Zoom-in-view of (c)

**Figure 11.** Localization and mapping results for the outdoor environment

| No | Method | Error mean x-axis (m) | Error variance x-axis (m) | Error mean y-axis (m) | Error variance y-axis (m) | Error mean z-axis (m) | Error variance z-axis (m) | Error mean $d$(m) | Error variance $d$(m) |
|---|---|---|---|---|---|---|---|---|---|
| 1 | TILAM | -0.0126 | 0.0898 | 0.0326 | 0.0105 | -0.0165 | 0.0018 | 0.0386 | 0.0904 |
| 2 | ICP based SLAM | -0.0125 | 0.1324 | -0.2016 | 0.0842 | -0.0781 | 0.0288 | 0.2166 | 0.1595 |

Table 1. Localization error comparison.

Moreover, there is a difference between the proposed TILAM and the ICP-based SLAM in terms of the time consumption, Table 2. According to Table 2, the time consumption of TIALM is divided into two parts: terrain inclination aided localization during the interval between any pair of two laser scans and ICP-based matching. Two ICP alignments were needed for the entire 15-meter path using the TILAM approach. The time consumption of ICP-based SLAM is also divided into two parts: odometry-based localization and ICP-based matching, where five ICP alignments were needed in order to achieve satisfying localization results. Although the terrain inclination aided localization took longer (0.09s in average) than the odometry-based localization (0.02s in average), the overall computational cost of TILAM (2.67s) was still 56% lower than the ICP-based SLAM (6.1s) because of a lower ICP alignment requirement with even better localization performance.

| No. | Test distance | Method | The consuming time for localization | The matching time | Total time |
|---|---|---|---|---|---|
| 1 | 15m | TILAM | 0.09 (s) x 3 times=0.27 (s) | 1.2 (s) x 2 times=2.4 (s) | 2.67 (s) |
| 2 | 15m | ICP based SLAM | 0.02 (s) x 5 times=0.1 (s) | 1.2 (s) x 5 times=6 (s) | 6.1 (s) |

Table 2. The computation time comparison.

In order to evaluate how the initial estimation error would affect the robot performance, the convergence periods are shown in Table 3 with different initial position estimations. Notice that the convergence period is less than 4s if the initial estimation error at each axis is less than 0.2 m. Moreover, it takes longer for the system to settle down as the initial estimation error increases. Hence, the proposed technique is sensitive to larger initial errors, although it only affects the localization accuracy at the beginning of the first scan interval and does not harm the performance afterwards. In the near future, a particle filter will be developed to improve the robustness of the proposed technique.

| No. | Robot speed (m/s) | Initial estimation error (m) | Convergence period (s) |
|---|---|---|---|
| 1 | 0.1 | (0.1,0.1,0.1) | 3.5 |
| 2 | 0.1 | (0.2,0.2,0.2) | 4 |
| 3 | 0.1 | (0.3,0.3,0.3) | 28 |
| 4 | 0.1 | (0.4,0.4,0.4) | 32 |

Table 3. Convergence periods with different initial estimation errors.

It is, therefore, concluded that the proposed terrain inclination aided localization and mapping algorithms can improve localization on 3D inclined terrains and efficiently build the global map of the surrounding area.

## 5. Future Work

In the near future, the proposed technique will be applied to achieve localization and mapping for a longer outdoor navigation in several, different environments. The accuracy and computational load will be further tested in a larger-scale area. In addition, in order to deal with the initial position errors, a particle filter will be developed to replace EKF in the future so as to achieve more robust performance.

## 6. Conclusions

This paper introduces a terrain inclination aided 3D localization and mapping technique for a mobile robot operating on outdoor inclined terrains. The experimental results validate the proposed technique which has capability to improve localization and mapping performance in terms of accuracy and computational cost. Future work will mainly focus on increasing the robustness of the algorithms and conducting more experiments in larger areas.

## 7. Acknowledgements

This research is currently supported by the NSF of China under NSFC grant no. 60905052 and by State Key Laboratory of Robotics and System (HIT) under SKLRS-2011-ZD-04. The suggestions from Leiming Guo and Liping Liu at the Harbin Institute of Technology were greatly appreciated during this research.